# Embodied Footprints: A Safety-guaranteed Collision-avoidance Model for Numerical Optimization-based Trajectory Planning

Bai Li, Youmin Zhang, *Fellow*, *IEEE*, Tantan Zhang, Tankut Acarman, Yakun Ouyang, Li Li, *Fellow*, *IEEE*, Hairong Dong, *Senior Member*, *IEEE*, and Dongpu Cao

*Abstract*—Optimization-based methods are commonly applied in autonomous driving trajectory planners, which transform the continuous-time trajectory planning problem into a finite nonlinear program with constraints imposed at finite collocation points. However, potential violations between adjacent collocation points can occur. To address this issue thoroughly, we propose a safety-guaranteed collision-avoidance model to mitigate collision risks within optimization-based trajectory planners. This model introduces an "embodied footprint", an enlarged representation of the vehicle's nominal footprint. If the embodied footprints do not collide with obstacles at finite collocation points, then the ego vehicle's nominal footprint is guaranteed to be collision-free at any of the infinite moments between adjacent collocation points. According to our theoretical analysis, we define the geometric size of an embodied footprint as a simple function of vehicle velocity and curvature. Particularly, we propose a trajectory optimizer with the embodied footprints that can theoretically set an appropriate number of collocation points prior to the optimization process. We conduct this research to enhance the foundation of optimization-based planners in robotics. Comparative simulations and field tests validate the completeness, solution speed, and solution quality of our proposal.

## I. INTRODUCTION

TRAJECTORY planning, a core module in an autonomous driving system, is designated to generate spatio-temporal curves that are kinematically feasible, collision-free, passenger-

Manuscript received February 10, 2023, revised July 22, 2023, September 10, 2023, accepted September 13, 2023. This work was supported by National Natural Science Foundation of China under Grant 62103139, National Key R&D Program of China under Grant 2022YFB2502905, Hejian Youth Talent Program of Hunan Province, China under Grant 2023RC3115, Fundamental Research Funds for the Central Universities under Grant 531118010509, and Natural Sciences & Engineering Research Council of Canada under Grant RGPIN-2023-05661. *(Corresponding author: Li Li)*

Bai Li, Tantan Zhang, and Yakun Ouyang are with the College of Mechanical and Vehicle Engineering, Hunan University, Changsha, China (e-mails: libai@zju.edu.cn, zhangtantan@hnu.edu.cn, yakun@hnu.edu.cn).

Youmin Zhang is with the Department of Mechanical, Industrial, and Aerospace Engineering, Concordia University, Montreal, QC H3G 1M8, Canada (e-mail: ymzhang@encs.concordia.ca).

Tankut Acarman is with the Department of Computer Engineering, Galatasaray University, Istanbul, Turkey (e-mail: tacarman@gsu.edu.tr).

Li Li is with the Department of Automation, BNRist, Tsinghua University, Beijing 100084, China (e-mail: li-li@tsinghua.edu.cn).

Hairong Dong is with the School of Electronics and Information Engineering, Beihang University, Beijing 100191, China (e-mail: hrdong@bjtu.edu.cn)

Dongpu Cao is with the School of Vehicle and Mobility, Tsinghua University, Beijing 100084, China (e-mail: dongpu.cao@uwaterloo.ca).

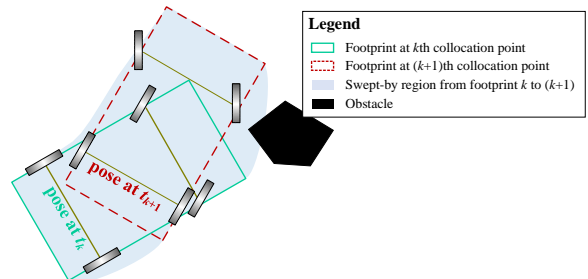

Fig. 1. Schematics on violations of collision-avoidance constraints between adjacent collocation points. Two discretized footprints are collision-free while collisions do occur between them.

friendly, and energy/time efficient [1,2]. The existing trajectory planners for autonomous vehicles are broadly classified into search-based, sampling-based, and numerical optimal control-based methods [3–6]. Search-/sampling-based planners seek a globally optimal path, while numerical optimal control-based planners aim for local optimality. This paper concentrates on numerical optimal control-based planning.

A numerical optimal control-based planner describes a nominal trajectory planning scheme as an optimal control problem (OCP). The OCP is discretized into a nonlinear program (NLP) problem before the NLP is solved by a gradient-based NLP solver [7]. This converts the continuous-time OCP into a problem with finite variables and constraints [8]. However, this conversion is imperfect because the time-continuous constraints in the original OCP are only enforced on finite collocation points, ignoring constraint satisfaction between adjacent collocation points. Increasing the density of collocation points can help, but it does not entirely solve the problem and results in high-dimensional NLPs that require extensive runtime. This constraint ignorance issue has been a common and silent drawback of numerical optimal control-based planners, particularly when constraints are highly non-convex [9–11].

Collision-avoidance constraints are widely acknowledged as the most complex type of constraint in a trajectory planning problem [12]. As depicted in Fig. 1, ignorance of collision-avoidance constraints between adjacent collocation points easily fails a planned trajectory [13]. This study aims to propose a constraint formulation method that theoretically guarantees safety between adjacent collocation points in an optimal control-based trajectory planner.

### A. Related Works

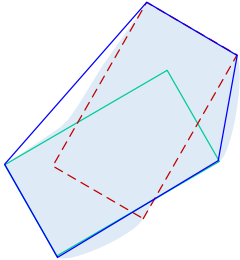

Fig. 2. Schematics on the usage of a convex hull to cover non-collocation point footprints. The edges of the convex hull are colored blue, which does not fully cover the swept-by region.

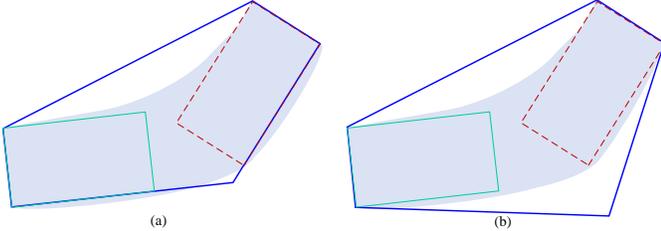

(a)  (b)

Fig. 3. Schematics on the usage of a polygonal region to cover the swept-by region between two adjacent vehicle footprints: (a) motion polygon proposed in [22]; (b) polygonal over-approximation method proposed in [23].

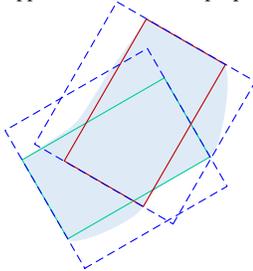

Fig. 4. Schematics on the usage of virtual protection frames to cover the swept-by region. Notably, the entire swept-by region is covered by two rectangles (i.e., the blue dashed boxes), which are derived by expanding the ego vehicle's footprints laterally.

This subsection reviews the previous studies that explored the violations of collision-avoidance constraints between adjacent collocation points or waypoints. For brevity, we refer to this violation as a *non-collocation error* throughout this paper.

Most prior works silently assume that non-collocation errors don't occur if 1) collocation points are *sufficiently* dense [14,15], 2) each polygonal obstacle is *sufficiently* dilated to form a buffer [16,17], or 3) a cost function with *sufficiently* large penalty weights is designed to keep the planned trajectory away from obstacles [18]. However, these assumptions lack a quantitative analysis of sufficiency, preventing them from becoming standardized knowledge.

The use of a convex hull to encompass the area swept by the vehicle footprint between two adjacent points has been considered [19,20]. Nevertheless, a convex hull does not guarantee full coverage of the swept region (Fig. 2). Schulman et al. [21] suggested expanding the convex hull, but the implementation was not detailed. They proposed a repulsive penalty in the cost function as a solution to non-collocation errors, but this does not fully eliminate collision risks. Scheuer and Fraichard [22] proposed a motion polygon to cover the swept-by region between adjacent collocation points, but part of the swept-by region is still out of the motion polygon (Fig. 3a). Additionally, their motion polygon cannot handle instances where adjacent poses overlap. Ghita and Kloetzer [23] proposed a polygonal over-approximation method, which is about building an expanded polygon with the intersection of two tangents to the swept-by region (Fig. 3b). However, that method is still too complex for a gradient-based NLP solver because the vertexes of the swept-by region are not easy to present. As seen from Fig. 3, the swept-by region is over-approximated in [22] or [23], leading to overcautious trajectories or even solution failures. To summarize, modeling the swept-by region as a compact polygon is inapplicable.

Contrary to compact polygon approaches, Li et al. [24] employed multiple polygons to cover the swept-by region. This strategy entails the interpolation of equidistant waypoints between two adjacent collocation points and the imposition of a collision-avoidance constraint at each interpolated waypoint. However, this strategy is still incomplete due to lingering minor collision risks. Zhang et al. [25] proposed a virtual protection frame (VPF) method, which covers the swept-by region by laterally expanded rectangles at each collocation point (Fig. 4). Unlike the compact polygon models, the VPF method ensures that the OCP dimension remains unchanged after addressing the non-collocation error. Nevertheless, the size of each expanded rectangle is established through trial and error, necessitating repeated iterations of the OCP solution process and consuming excessive runtime. More critically, determining the expanded rectangle's size via trial-and-error risks failure of the iterative solution process, particularly in a tiny environment (this is demonstrated through comparative simulations in Section V). Similar to VPF, methods like the adaptive mesh refinement by Yershov and Frazzoli [26], and the moving finite element method by Chen et al. [27] also exist. In [26], the trajectory resolution is enhanced by continually adjusting the discretized mesh grids, while [27] persistently shifts the non-uniform collocation points until the non-collocation error is minimized. However, both [26] and [27] operate on a trial-and-error basis, making them challenging for quick execution.

In summary, no previous study has satisfactorily addressed the non-collocation error issue considering both completeness and solution speed.

*B. Motivations and Contributions*

Upon reviewing previous studies, we noted that 1) increasing the density of collocation points would not completely address the issue of concern, and 2) the compact polygon-based methods are excessively complex. The VPF method is promising because it addresses the non-collocation error issue by operating only on collocation points. However, the VPF method is imperfect because it calls for an iterative planning process to determine the expansion buffers, which consumes runtime and easily makes the planner incomplete. In response to the limitations of the VPF method, this work aims to theoretically define expansion buffers, eliminating the need for time-consuming iterative computation.

The contribution of this work lies in the theoretical development of a collision-avoidance constraint model, referred to as the embodied footprint model, which ensures complete safety between adjacent collocation points in using a numerical optimal control-based planner. Specifically, the vehicle footprint at each collocation point is expanded longitudinally and laterally by buffers that fully cover the swept-by region between adjacent collocation points. These buffers, directly related to vehicle speed and curvature, are treated as decision variables along with other state/control variables within the OCP. This gives a numerical optimal control-based planner enough flexibility to find a safe trajectory without resorting to iterative computation.

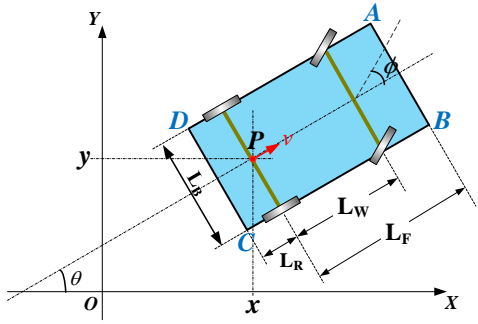

Fig. 5. Schematics on vehicle kinematics and geometrics.

### C. Organization

In the rest of this paper, Section II states the concerned problem in a generic trajectory planning background. Section III proposes a safe collision-avoidance constraint modeling method. Section IV introduces a trajectory planner embedded with the modeling method introduced in Section III. Section V presents and discusses the simulation and experimental results. Finally, Section VI concludes the paper.

## II. TRAJECTORY PLANNING PROBLEM STATEMENT

This section outlines the trajectory planning problem. A generic formulation of a trajectory planning task is provided in the form of an OCP, which is then discretized into an NLP problem.

### A. OCP Formulation

A trajectory planning task is cast into the following OCP:

$$\begin{aligned}
&\underset{\mathbf{z}(t),\, \mathbf{u}(t),\, T}{\text{minimize}} \; J, \\
&\text{s.t., } \dot{\mathbf{z}}(t) = f(\mathbf{z}(t),\, \mathbf{u}(t)),\; t \in [0, T]; \\
&\quad \underline{\mathbf{z}} \le \mathbf{z}(t) \le \overline{\mathbf{z}},\; \underline{\mathbf{u}} \le \mathbf{u}(t) \le \overline{\mathbf{u}},\; t \in [0, T]; \\
&\quad \mathbf{z}(0) = \mathbf{z}_{\text{init}},\; \mathbf{u}(0) = \mathbf{u}_{\text{init}}; \\
&\quad \mathbf{z}(T) = \mathbf{z}_{\text{end}},\; \mathbf{u}(T) = \mathbf{u}_{\text{end}}; \\
&\quad fp(\mathbf{z}(t)) \subset \Upsilon_{\text{FREE}},\; t \in [0, T].
\end{aligned} \quad (1)$$

In this formulation, $\mathbf{z} \in \mathbb{R}^{n_z}$ represents the vehicle state profiles, $\mathbf{u} \in \mathbb{R}^{n_u}$ denotes the control profiles, and $T$ represents the unfixed planning horizon length.

We assume that the single-track bicycle model [7] is able to describe the kinematic constraints $\dot{\mathbf{z}}(t) = f(\mathbf{z}(t),\, \mathbf{u}(t))$:

$$\frac{d}{dt}\begin{bmatrix} x(t) \\ y(t) \\ v(t) \\ \phi(t) \\ \theta(t) \end{bmatrix} = \begin{bmatrix} v(t) \cdot \cos\theta(t) \\ v(t) \cdot \sin\theta(t) \\ a(t) \\ \omega(t) \\ v(t) \cdot \tan\phi(t)/L_W \end{bmatrix}, \quad (2)$$

$t \in [0, T]$.

Herein, $(x, y)$ represents the coordinate value of the midpoint along the rear-wheel axle of the ego vehicle (Fig. 5), $\theta$ refers to the orientation angle, $v$ is the vehicle velocity, $\phi$ represents the steering angle, $a$ is the acceleration, $\omega$ denotes the angular velocity of the steering angle, and $L_W$ is the wheelbase. The other geometric parameters marked in Fig. 5 include $L_F$ (front hang length plus wheelbase), $L_R$ (rear hang length), and $L_B$ (width).

$[\underline{\mathbf{z}}, \overline{\mathbf{z}}]$ and $[\underline{\mathbf{u}}, \overline{\mathbf{u}}]$ denote the allowable intervals for $\mathbf{z}(t)$ and $\mathbf{u}(t)$, respectively, that is,

$$\begin{bmatrix} a_{\min} \\ 0 \\ -\Omega_{\max} \\ -\Phi_{\max} \end{bmatrix} \le \begin{bmatrix} a(t) \\ v(t) \\ \omega(t) \\ \phi(t) \end{bmatrix} \le \begin{bmatrix} a_{\max} \\ v_{\max} \\ \Omega_{\max} \\ \Phi_{\max} \end{bmatrix}, \quad (3)$$

$t \in [0, T]$.

Here, $v(t) \ge 0$ assumes that backward maneuvers are ignored in this present work. $a_{\min}$, $a_{\max}$, $v_{\max}$, $\Omega_{\max}$, and $\Phi_{\max}$ are boundary parameters.

$\mathbf{z}_{\text{init}}$ and $\mathbf{u}_{\text{init}}$ denote the initial-moment values of $\mathbf{z}(t)$ and $\mathbf{u}(t)$. $\mathbf{z}_{\text{end}}$ and $\mathbf{u}_{\text{end}}$ denote their terminal-moment values, respectively.

$fp(\cdot): \mathbb{R}^{n_z} \to \mathbb{R}^2$ is a mapping from the vehicle's state profile $\mathbf{z}(t)$ to the footprint. Thus, $\Upsilon_{\text{FREE}}$ denotes the free space in the 2D environment while $fp(\mathbf{z}(t)) \subset \Upsilon_{\text{FREE}}$ represents the collision-avoidance constraints. This work assumes that $\Upsilon_{\text{FREE}}$ is static, i.e., no moving obstacles are considered. For future usage, let us define the four vertexes of the ego vehicle's footprint as $A = (x_A, y_A)$, $B = (x_B, y_B)$, $C = (x_C, y_C)$, and $D = (x_D, y_D)$ (Fig. 5):

$$\begin{aligned}
x_A(t) &= x(t) + L_F \cdot \cos\theta(t) - 0.5 L_B \cdot \sin\theta(t), \\
y_A(t) &= y(t) + L_F \cdot \sin\theta(t) + 0.5 L_B \cdot \cos\theta(t), \\
x_B(t) &= x(t) + L_F \cdot \cos\theta(t) + 0.5 L_B \cdot \sin\theta(t), \\
y_B(t) &= y(t) + L_F \cdot \sin\theta(t) - 0.5 L_B \cdot \cos\theta(t), \\
x_C(t) &= x(t) - L_R \cdot \cos\theta(t) + 0.5 L_B \cdot \sin\theta(t), \\
y_C(t) &= y(t) - L_R \cdot \sin\theta(t) - 0.5 L_B \cdot \cos\theta(t), \\
x_D(t) &= x(t) - L_R \cdot \cos\theta(t) - 0.5 L_B \cdot \sin\theta(t), \\
y_D(t) &= y(t) - L_R \cdot \sin\theta(t) + 0.5 L_B \cdot \cos\theta(t),
\end{aligned} \quad (4)$$

$t \in [0, T]$.

The concrete expression of $fp(\mathbf{z}(t)) \subset \Upsilon_{\text{FREE}}$ is presented later in Section III.E.

### B. NLP Formulation

The previous subsection casts the trajectory planning scheme as an OCP. This subsection briefly outlines the principle for discretizing it into an NLP. For simplicity, the OCP constructed in the previous subsection is abstracted as

$$\begin{aligned}
&\underset{\mathbf{z}(t),\, \mathbf{u}(t),\, T}{\text{minimize}} \; J, \\
&\text{s.t., } g(\mathbf{z}(t), \dot{\mathbf{z}}(t), \mathbf{u}(t)) \le 0,\; t \in [0, T].
\end{aligned} \quad (5)$$

Herein, $g \le 0$ represents all the constraints in inequality and equality forms.

Discretizing (5) into an NLP is about sampling finite moments along the time dimension $t \in [0, T]$ such that each $\mathbf{z}(t)$ or $\mathbf{u}(t)$ could be represented by finite collocation points. In this manner, the constraints $g \le 0$ apply only to finite collocation points rather than the entire time domain $[0, T]$. Suppose that $(N_{\text{fe}} + 1)$ moments are sampled from 0 to $T$, which are collected in a set $\{t_i \mid i = 0, ..., N_{\text{fe}}\}$ with

$$0 = t_0 < t_1 < t_2 < ... < t_{N_{fe}} = T. \quad (6)$$

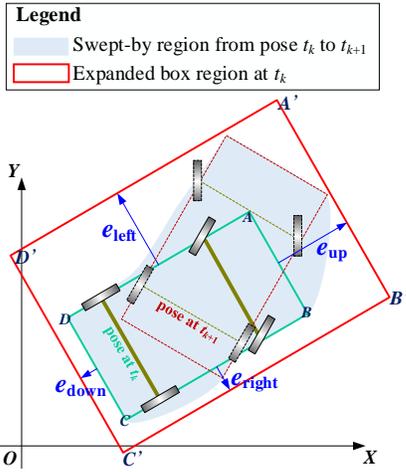

Fig. 6. Schematics on embodied box $A'B'C'D'$ that covers the swept-by region from $t_k$ to $t_{k+1}$.

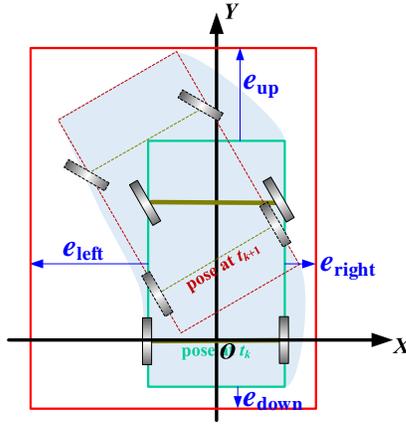

Fig. 7. A normalized angle of review of Fig. 6 to facilitate theoretical analysis.

Each $\mathbf{z}(t)$ is represented by ($N_{fe} + 1$) collocation points, which are denoted by $\{z_i | i = 0,...,N_{fe}\}$. These collocation points are used to construct an infinite-dimensional variable $\mathbf{z}(t)$, which might be a piecewise constant, linear, or polynomial function. Similarly, $\mathbf{u}(t)$ is represented by $\{u_i | i = 0,...,N_{fe}\}$. Thus, the original OCP is discretized into the following NLP:

$$\begin{aligned} &\min_{z_i, u_i, t_i} J(z_i, u_i, t_i), \\ &\text{s.t., } g(z_i, u_i, t_i) \leq 0, \ i = 0,...,N_{fe}. \end{aligned} \quad (7)$$

## III. EMBODIED FOOTPRINT-BASED COLLISION AVOIDANCE

Nominally, one uses an NLP solver to solve (7), thereby deriving a numerical solution to the original OCP (5). However, the derived trajectory is not guaranteed to be safe because collision-avoidance constraints are only imposed at finite moments $\{t_i\}$. To resolve this issue thoroughly, we continue to impose the collision-avoidance constraints only on finite collocation points, but we expect each of them to be "responsible", ensuring the entire time horizon $t \in [0,T]$ is completely safe. This necessitates changes in the collision-avoidance constraints.

This section introduces how to model safe collision-avoidance constraints in the NLP formulation (7). As a basic step, the modeling task is identified in Section III.A before details are provided in the next few subsections.

### A. Collision-avoidance Constraint Modeling Task Statement

This subsection states the safety-oriented collision-avoidance constraint modeling. Since the collision-avoidance constraints are used in the formulated NLP, we have to impose a finite number of collision-avoidance constraints, but we expect that each collision-avoidance constraint is modeled in a way that ensures the ego vehicle is safe within a time interval. Specifically, we expand the ego vehicle's footprint laterally and longitudinally to define an "embodied box" and require that the swept-by region from $t_k$ to $t_{k+1}$ is fully covered by an embodied box over the ego vehicle's footprint at $t_k$. As shown in Fig. 6, each embodied box is aligned with the vehicle footprint, and the geometric size is determined by four variables at $t_k$, which are denoted as $e_{\text{left}}(t_k)$, $e_{\text{right}}(t_k)$, $e_{\text{up}}(t_k)$, and $e_{\text{down}}(t_k)$. The central challenge in this section is defining the four variables at $t_k$ in a way that thoroughly eliminates collision risks between adjacent collocation points $t_k$ and $t_{k+1}$.

Without loss of generality, we rotate the vehicle in Fig. 6 to an axially aligned pose (see Fig. 7) for the convenience of analysis, i.e., $[x(t_k), y(t_k), \theta(t_k)] = [0,0,\pi/2]$. Owing to this rotation, the four variables can be easily presented as

$$\begin{aligned} e_{\text{left}}(t_k) &= -\frac{L_B}{2} - x_{\min}, \\ e_{\text{right}}(t_k) &= x_{\max} - \frac{L_B}{2}, \\ e_{\text{up}}(t_k) &= y_{\max} - L_F, \\ e_{\text{down}}(t_k) &= -L_R - y_{\min}, \end{aligned} \quad (8)$$

where $x_{\min}$, $x_{\max}$, $y_{\min}$, and $y_{\max}$ denote the lower/upper bounds of the swept-by region, respectively. The determinations of $x_{\min}$, $x_{\max}$, $y_{\min}$, and $y_{\max}$ naturally decide $e_{\text{left}}(t_k)$, $e_{\text{right}}(t_k)$, $e_{\text{up}}(t_k)$, and $e_{\text{down}}(t_k)$. Therefore, the core problem in modeling the safe collision-avoidance constraints is how to determine $x_{\min}$, $x_{\max}$, $y_{\min}$, and $y_{\max}$. Our proposed idea is presented in Sections III.B and III.C.

### B. Determination of $y_{\min}$

This subsection introduces how to identify $y_{\min}(t_k)$, which denotes the minimal coordinate value of the swept-by region from $t_k$ to $t_{k+1}$. For brevity, we will refer to $y_{\min}(t_k)$ as $y_{\min}$.

Nominally, the shape of the swept-by region is determined by control variables $\mathbf{u}(t)$ during the time interval $[t_k, t_{k+1}]$. If $\mathbf{u}(t)$ is not constant during $[t_k, t_{k+1}]$, then the swept-by region would be too complicated to analyze. We assume that the state variables $\mathbf{z}(t)$ are constant during $[t_k, t_{k+1}]$:

***Assumption 1***: Any a $\mathbf{z}(t)$ or $\mathbf{u}(t)$ in the concerned OCP (1) is a piecewise constant function.

If $\mathbf{u}(t)$ is assumed to be piecewise constant on a time interval, then $\mathbf{z}(t)$ may not be piecewise constant on that interval. Assuming both $\mathbf{z}(t)$ and $\mathbf{u}(t)$ are piecewise constant may violate the kinematic constraints (2) in the concerned OCP. Despite this, Assumption 1 is a common practice in numerical optimization that causes minor errors if the collocation points are set densely. Discretizing OCP (1) with this assumption is known as the explicit first-order Runge–Kutta method. We must uphold this assumption, as without it, analyzing the shape of a swept-by region is extremely difficult.

Assumption 1 yields that the steering angle variable $\phi(t)$ remains constant during $t \in [t_k, t_{k+1}]$, thus the path segment of the rear-axle midpoint located at $(x(t), y(t))$ is circular during $t \in [t_k, t_{k+1}]$. Similarly, the velocity variable $v(t)$ is constant during $t \in [t_k, t_{k+1}]$. For simplicity, let us rewrite the constant steering angle as $\phi_k$ and the constant velocity as $v_k$. According to (2) and the Newton–Leibniz formula, the orientation angle $\theta(t)$ is determined as

$$\theta(t) = \theta(t_k) + \int_{\tau=t_k}^{t} \frac{v(t) \cdot \tan \phi(t)}{L_W} d\tau$$
$$= \frac{\pi}{2} + \frac{v_k \cdot \tan \phi_k}{L_W} \cdot (t - t_k), \ t \in [t_k, t_{k+1}]. \quad (9)$$

To further simplify the presentation, let us introduce a curvature variable $\kappa(t) \equiv \tan \phi(t)/L_W$ and set that $t_k = 0$ and $t_{k+1} = \Delta T$, then we have

$$\theta(t) = \frac{\pi}{2} + v_k \cdot \kappa_k \cdot t, \ t \in [0, \Delta T], \quad (10a)$$

where $\kappa_k = \tan \phi_k / L_W$. Similarly, $x(t)$ and $y(t)$ are defined as

$$x(t) = \frac{\cos(v_k \cdot \kappa_k \cdot t) - 1}{\kappa_k}, \ t \in [0, \Delta T], \quad (10b)$$

and

$$y(t) = \frac{\sin(v_k \cdot \kappa_k \cdot t)}{\kappa_k}, \ t \in [0, \Delta T]. \quad (10c)$$

Substituting (10) into (4) yields the trajectories of the ego vehicle's four vertex points:

$$x_A(t) = \frac{\cos(v_k \cdot \kappa_k \cdot t) - 1}{\kappa_k} - L_F \cdot \sin(v_k \cdot \kappa_k \cdot t) - \frac{L_B}{2} \cdot \cos(v_k \cdot \kappa_k \cdot t),$$

$$y_A(t) = \frac{\sin(v_k \cdot \kappa_k \cdot t)}{\kappa_k} + L_F \cdot \cos(v_k \cdot \kappa_k \cdot t) - \frac{L_B}{2} \cdot \sin(v_k \cdot \kappa_k \cdot t),$$

$$x_B(t) = \frac{\cos(v_k \cdot \kappa_k \cdot t) - 1}{\kappa_k} - L_F \cdot \sin(v_k \cdot \kappa_k \cdot t) + \frac{L_B}{2} \cdot \cos(v_k \cdot \kappa_k \cdot t),$$

$$y_B(t) = \frac{\sin(v_k \cdot \kappa_k \cdot t)}{\kappa_k} + L_F \cdot \cos(v_k \cdot \kappa_k \cdot t) + \frac{L_B}{2} \cdot \sin(v_k \cdot \kappa_k \cdot t),$$

$$x_C(t) = \frac{\cos(v_k \cdot \kappa_k \cdot t) - 1}{\kappa_k} + L_R \cdot \sin(v_k \cdot \kappa_k \cdot t) + \frac{L_B}{2} \cdot \cos(v_k \cdot \kappa_k \cdot t),$$

$$y_C(t) = \frac{\sin(v_k \cdot \kappa_k \cdot t)}{\kappa_k} - L_R \cdot \cos(v_k \cdot \kappa_k \cdot t) + \frac{L_B}{2} \cdot \sin(v_k \cdot \kappa_k \cdot t),$$

$$x_D(t) = \frac{\cos(v_k \cdot \kappa_k \cdot t) - 1}{\kappa_k} + L_R \cdot \sin(v_k \cdot \kappa_k \cdot t) - \frac{L_B}{2} \cdot \cos(v_k \cdot \kappa_k \cdot t),$$

$$y_D(t) = \frac{\sin(v_k \cdot \kappa_k \cdot t)}{\kappa_k} - L_R \cdot \cos(v_k \cdot \kappa_k \cdot t) - \frac{L_B}{2} \cdot \sin(v_k \cdot \kappa_k \cdot t),$$

$$t \in [0, \Delta T]. \quad (11)$$

Connecting the four vertexes $A$, $B$, $C$, and $D$ forms a rectangular footprint, which moves throughout $t \in [0, \Delta T]$ to create a swept-by region. Obviously, the boundaries of the swept-by region are determined by vertices rather than edges of the footprint, thus

$$x_{\min} = \min\{x_A(t), x_B(t), x_C(t), x_D(t), \forall t \in [0, \Delta T]\}, \quad (12a)$$

$$x_{\max} = \max\{x_A(t), x_B(t), x_C(t), x_D(t), \forall t \in [0, \Delta T]\}, \quad (12b)$$

$$y_{\min} = \min\{y_A(t), y_B(t), y_C(t), y_D(t), \forall t \in [0, \Delta T]\}, \quad (12c)$$

$$y_{\max} = \max\{y_A(t), y_B(t), y_C(t), y_D(t), \forall t \in [0, \Delta T]\}. \quad (12d)$$

According to the definitions of $y_A(t), y_B(t), y_C(t),$ and $y_D(t)$ in (11), one can safely state that $y_A(t) \geq y_D(t)$ and $y_B(t) \geq y_C(t)$ for any $t$ provided that

$$|v_k \cdot \kappa_k \cdot t| \leq \frac{\pi}{2}, \ \forall t \in [0, \Delta T]. \quad (13)$$

Given that (13) should be satisfied for any $t \in [0, \Delta T]$, then

$$|v_k \cdot \kappa_k \cdot \Delta T| \leq \frac{\pi}{2}. \quad (14)$$

**Assumption 2**: $v_k$, $\kappa_k$, and $\Delta T$ satisfy the inequality (14).

Assumption 2 ensures that $y_A(t) \geq y_D(t)$ and $y_B(t) \geq y_C(t)$ for any $t$ within $[0, \Delta T]$. Therefore, $y_{\min}$ would be the smaller one between $\min\{y_C(t)\}$ and $\min\{y_D(t)\}$ for $\forall t \in [0, \Delta T]$. If Assumption 2 is not held, one needs to also consider the complex case in which $y_{\min}$ involves vertex $A$ or $B$.

Recall that negative velocity is not considered in this present work as mentioned in (3), thus we state the following:

**Assumption 3**: $v_k \geq 0$.

With Assumption 3 at hand, the subsequent analyses are divided into two branches as per the sign of $\kappa_k$.

**Condition 1:** $\kappa_k \geq 0$

When $\kappa_k$ is non-negative, one has $y_D(t) \leq y_C(t)$. Thus, $y_{\min}$ is the extremum of $y_D(t)$ on $t \in [0, \Delta T]$. Let us rewrite $y_D(t)$ in the following form:

$$y_D(t) = (\frac{1}{\kappa_k} - \frac{L_B}{2}) \cdot \sin(v_k \cdot \kappa_k \cdot t) - L_R \cdot \cos(v_k \cdot \kappa_k \cdot t)$$
$$= M \cdot \sin(v_k \cdot \kappa_k \cdot t) - N \cdot \cos(v_k \cdot \kappa_k \cdot t) \quad (15)$$
$$= \sqrt{M^2 + N^2} \cdot \sin(v_k \cdot \kappa_k \cdot t - \alpha),$$

where $M \equiv \frac{1}{\kappa_k} - \frac{L_B}{2}$, $N \equiv L_R$, and

$$\alpha = \arccos(\frac{M}{\sqrt{M^2 + N^2}}). \quad (16)$$

Herein, N is positive while the sign of M is pending. Let us discuss the sign of M.

If $M < 0$, then $M/\sqrt{M^2 + N^2}$ is negative, thus (16) yields that $\alpha \in [\pi/2, \pi]$. Let us analyze the monotonicity of $y_D(t)$ via its derivative $y_D'(t)$:

$$y_D'(t) = v_k \cdot \kappa_k \cdot \sqrt{M^2 + N^2} \cdot \cos(v_k \cdot \kappa_k \cdot t - \alpha).$$

Given that $v_k \cdot \kappa_k \cdot \sqrt{M^2 + N^2} > 0$, the sign of $y_D'(t)$ would be determined by $\cos(v_k \cdot \kappa_k \cdot t - \alpha)$. Obviously, $y_D'(0) \leq 0$, which means $y_D(t)$ has a descending trend at $t = 0^+$. We expect that the entire domain $[0, \Delta T]$ is a monotonic decreasing interval. If so, the minimal extremum value is $y_D(\Delta T)$. The reason why we expect $[0, \Delta T]$ to be monotonic is given as follows. If $[0, \Delta T]$ is not a monotonic decreasing interval, then the extremum is achieved at some $t^* \in [0, \Delta T]$ that satisfies

$$\sin(v_k \cdot \kappa_k \cdot t^* - \alpha) = -1,$$

which renders the following side effects: 1) the extremum is unrelated to $v_k$, thus making $y_{\min}$ overcautious; and 2) the extremum is a rather complex function of $\kappa_k$.

To make $[0, \Delta T]$ a monotonic decreasing interval, one needs to ensure that $\cos(v_k \cdot \kappa_k \cdot t - \alpha)$ remains negative throughout $[0, \Delta T]$, which yields $\cos(v_k \cdot \kappa_k \cdot \Delta T - \alpha) \leq 0$, i.e.,

$$v_k \cdot \kappa_k \cdot \Delta T - \alpha \leq -\frac{\pi}{2}. \quad (17)$$

This inequality is further described as

$$0 < v_k \cdot \kappa_k \cdot \Delta T \leq \arctan(-\frac{\frac{1}{\kappa_k} - \frac{L_B}{2}}{L_R}). \quad (18)$$

If (18) holds, then the extremum of $y_D(t)$ is

$$y_D(\Delta T) = (\frac{1}{\kappa_k} - \frac{L_B}{2}) \cdot \sin(v_k \cdot \kappa_k \cdot \Delta T) - L_R \cdot \cos(v_k \cdot \kappa_k \cdot \Delta T). \quad (19)$$

Nominally, one can state $y_{min} = y_D(\Delta T)$, but (19) is still not showing a simple enough relationship among $y_{min}$, $v_k$, and $\kappa_k$. Therefore, we simplify (19) further via inequality amplification skills in mathematics. According to the McLaughlin formula, one has

$$\sin(v_k \cdot \kappa_k \cdot \Delta T) \leq v_k \cdot \kappa_k \cdot \Delta T, \quad (20a)$$

and

$$\cos(v_k \cdot \kappa_k \cdot \Delta T) \leq 1. \quad (20b)$$

Then, we can provide a simplified lower bound for $y_D(\Delta T)$:

$$(\frac{1}{\kappa_k} - \frac{L_B}{2}) \cdot v_k \cdot \kappa_k \cdot \Delta T - L_R \leq y_D(\Delta T). \quad (21)$$

Accordingly, one has

$$y_{min} = \frac{L_B}{2} \cdot \kappa_k \cdot s_k - s_k - L_R, \quad (22)$$

wherein $s_k \equiv v_k \cdot \Delta T$ is deployed to simplify the presentation.

The aforementioned analysis assumes that $M < 0$. Symmetrically, if $M \geq 0$, then $\alpha \in [0, \pi/2]$ and $[0, \Delta T]$ is definitely a monotonic increasing interval. Therefore, the extremum value is $y_D(0)$, i.e., $y_{min} = -L_R$.

**Condition 2:** $\kappa_k < 0$

In dealing with $\kappa_k < 0$, we introduce a temporary variable $\kappa_k^* = -\kappa_k$. Since $\kappa_k^* > 0$, the analysis in **Condition 1** could be repeated. The concrete details are similar, thus we omit them.

Summarizing the analyses for $\kappa_k \geq 0$ and $\kappa_k < 0$ yields that

1° When $\frac{1}{|\kappa_k|} - \frac{L_B}{2} \leq 0$,

$$y_{min} = \frac{L_B}{2} \cdot |\kappa_k| \cdot s_k - s_k - L_R, \quad (23a)$$

which is associated with prerequisite

$$|\kappa_k| \cdot s_k \leq \arctan(-\frac{\frac{1}{|\kappa_k|} - \frac{L_B}{2}}{L_R}). \quad (23b)$$

2° When $\frac{1}{|\kappa_k|} - \frac{L_B}{2} > 0$,

$$y_{min} = -L_R, \quad (23c)$$

which is associated with prerequisite

$$|\kappa_k| \cdot s_k \leq \frac{\pi}{2}. \quad (23d)$$

*Assumption 4*: $2L_W > L_B \cdot \tan\phi_{max}$.

Empirically, Assumption 4 holds for a passenger vehicle, which means

$$\frac{1}{|\kappa_k|} - \frac{L_B}{2} = \left|\frac{L_W}{\tan\phi_k}\right| - \frac{L_B}{2} \geq \frac{L_W}{\tan\phi_{max}} - \frac{L_B}{2} > 0. \quad (24)$$

Therefore, the branch under $1/|\kappa_k| - L_B/2 \leq 0$ is discarded under Assumption 4. The cases that involve $1/|\kappa_k| - L_B/2 \leq 0$ will not be considered in the remainder of this section either.

The final conclusion of this subsection is that $y_{min} = -L_R$ under Assumptions 1–4.

*C. Determinations of $y_{max}$, $x_{min}$, and $x_{max}$*

This subsection introduces the definitions of $y_{max}$, $x_{min}$, and $x_{max}$. The analyses are similar to those in Section III.B, thus details are omitted.

$y_{max}$ is written as

$$y_{max} = s_k + \frac{L_B}{2} \cdot |\kappa_k| \cdot s_k + L_F, \quad (25a)$$

together with the following prerequisite:

$$s_k \cdot |\kappa_k| \leq \arctan(\frac{\frac{1}{|\kappa_k|} + \frac{L_B}{2}}{L_F}). \quad (25b)$$

$x_{min}$ is written as

$$x_{min} = -\frac{L_B}{2} - \max\left\{-L_R \cdot \kappa_k \cdot s_k, \; (L_F + \frac{s_k}{2}) \cdot \kappa_k \cdot s_k\right\}, \quad (26a)$$

together with

$$s_k \cdot |\kappa_k| \leq \arctan(\frac{L_R}{\frac{1}{|\kappa_k|} + \frac{L_B}{2}}). \quad (26b)$$

$x_{max}$ is written as

$$x_{max} = \frac{L_B}{2} + \max\left\{L_R \cdot \kappa_k \cdot s_k, \; -(L_F + \frac{s_k}{2}) \cdot \kappa_k \cdot s_k\right\}, \quad (27a)$$

together with

$$s_k \cdot |\kappa_k| \leq \arctan(\frac{L_R}{\frac{1}{|\kappa_k|} + \frac{L_B}{2}}). \quad (27b)$$

*D. Definition of an Embodied Box*

Sections III.B and III.C have defined $y_{min}$, $y_{max}$, $x_{min}$, and $x_{max}$. This subsection further forms an embodied box based on the aforementioned definitions. Eq. (8) yields that

$$(-x_{min}) - e_{left}(t_k) = \frac{L_B}{2},$$

$$x_{max} - e_{right}(t_k) = \frac{L_B}{2}, \quad (28)$$

$$(-y_{min}) - e_{down}(t_k) = L_R,$$

$$y_{max} - e_{up}(t_k) = L_F.$$

Eq. (28), Section III.B, and III.C yield that

$$e_{left}(t_k) = \max\left\{-L_R \cdot \kappa_k \cdot s_k, \; (L_F + \frac{s_k}{2}) \cdot \kappa_k \cdot s_k\right\},$$

$$e_{right}(t_k) = \max\left\{L_R \cdot \kappa_k \cdot s_k, \; -(L_F + \frac{s_k}{2}) \cdot \kappa_k \cdot s_k\right\}, \quad (29)$$

$$e_{up}(t_k) = s_k + \frac{L_B}{2} \cdot |\kappa_k| \cdot s_k,$$

$$e_{down}(t_k) = 0,$$

subject to the intersection of prerequisites (23d), (25b), (26b), and (27b). Since (26b) and (27b) are identical, one only needs to consider (23d), (25b), and (26b).

Eq. (25b) is rewritten as

$$\tan\left(s_k \cdot |\kappa_k|\right) \leq \frac{\frac{1}{|\kappa_k|} + \frac{L_B}{2}}{L_F}, \tag{30a}$$

that is,

$$|\kappa_k| \cdot L_F \cdot \tan\left(s_k \cdot |\kappa_k|\right) \leq 1 + \frac{L_B}{2} \cdot |\kappa_k|. \tag{30b}$$

Similarly, (26b) is rewritten as

$$\left(1 + \frac{L_B}{2} \cdot |\kappa_k|\right) \cdot \tan\left(s_k \cdot |\kappa_k|\right) \leq L_R \cdot |\kappa_k|. \tag{31}$$

Based on the aforementioned analyses, the four vertexes of the ego vehicle at time instance $t_k$ are expanded as the rectangle $A'B'C'D'$ shown in Fig. 6, where $e_{\text{left}}(t_k)$, $e_{\text{right}}(t_k)$, $e_{\text{up}}(t_k)$, and $e_{\text{down}}(t_k)$ are defined in (29). Simultaneously, prerequisites (23d), (30b), and (31) should hold.

According to (29), the embodied box becomes large when $|\kappa_k|$ and/or $s_k$ are large, which is intuitively reasonable. Prerequisites (23d), (30b), and (31) are inherently setting an upper bound for $(t_{k+1} - t_k)$, i.e., the duration between adjacent collocation points.

### E. Formulation of Safe Collision-avoidance Constraints

This subsection presents the safety-oriented collision-avoidance constraints with the defined embodied box at the collocation point $t_k$.

The four vertices of the embodied box are defined as $A' = (\hat{x}_A, \hat{y}_A)$, $B' = (\hat{x}_B, \hat{y}_B)$, $C' = (\hat{x}_C, \hat{y}_C)$, and $D' = (\hat{x}_D, \hat{y}_D)$:

$$\hat{x}_A(t_k) = x(t_k) + \left(L_F + e_{\text{up}}(t_k)\right) \cdot \cos\theta(t_k) - \left(\frac{L_B}{2} + e_{\text{left}}(t_k)\right) \cdot \sin\theta(t_k),$$

$$\hat{y}_A(t_k) = y(t_k) + \left(L_F + e_{\text{up}}(t_k)\right) \cdot \sin\theta(t_k) + \left(\frac{L_B}{2} + e_{\text{left}}(t_k)\right) \cdot \cos\theta(t_k),$$

$$\hat{x}_B(t_k) = x(t_k) + \left(L_F + e_{\text{up}}(t_k)\right) \cdot \cos\theta(t_k) + \left(\frac{L_B}{2} + e_{\text{right}}(t_k)\right) \cdot \sin\theta(t_k),$$

$$\hat{y}_B(t_k) = y(t_k) + \left(L_F + e_{\text{up}}(t_k)\right) \cdot \sin\theta(t_k) - \left(\frac{L_B}{2} + e_{\text{right}}(t_k)\right) \cdot \cos\theta(t_k),$$

$$\hat{x}_C(t_k) = x(t_k) - \left(L_R + e_{\text{down}}(t_k)\right) \cdot \cos\theta(t_k) + \left(\frac{L_B}{2} + e_{\text{right}}(t_k)\right) \cdot \sin\theta(t_k),$$

$$\hat{y}_C(t_k) = y(t_k) - \left(L_R + e_{\text{down}}(t_k)\right) \cdot \sin\theta(t_k) - \left(\frac{L_B}{2} + e_{\text{right}}(t_k)\right) \cdot \cos\theta(t_k),$$

$$\hat{x}_D(t_k) = x(t_k) - \left(L_R + e_{\text{down}}(t_k)\right) \cdot \cos\theta(t_k) - \left(\frac{L_B}{2} + e_{\text{left}}(t_k)\right) \cdot \sin\theta(t_k),$$

$$\hat{y}_D(t_k) = y(t_k) - \left(L_R + e_{\text{down}}(t_k)\right) \cdot \sin\theta(t_k) + \left(\frac{L_B}{2} + e_{\text{left}}(t_k)\right) \cdot \cos\theta(t_k). \tag{32}$$

It is required that the embodied box $A'B'C'D'$ does not overlap with obstacles at ($N_{fe}-1$) moments $\{t_k | k=1,...,N_{fe}-1\}$. Before ending this section, we briefly present the principle to describe $fp(\mathbf{z}(t)) \subset \Upsilon_{\text{FREE}}$ as algebraic inequalities via a triangle-area criterion introduced in [7].

A collision between the rectangular embodied box $A'B'C'D'$ and each obstacle should be avoided at the ($N_{fe}-1$) collocation points. Without loss of generality, the collision-avoidance constraint between the $j$th obstacle and the embodied box $A'B'C'D'$ at $t_k$ is examined. If the $j$th obstacle has $N_j$ vertices denoted as $V_{j1},...,V_{jN_j}$, a collision begins when a vertex of the obstacle hits the ego vehicle's embodied box $A'B'C'D'$ or vice versa, a vertex of the embodied box hits in the obstacle. Hence, collisions will not occur if 1) vertexes $A'$, $B'$, $C'$, and $D'$ are always located out of the $j$th obstacle, and 2) each $V_{jk}$ ($k=1,...,N_j$) always locate out of $A'B'C'D'$.

The generic constraint that a point $Q = (x_Q, y_Q)$ locates outside a convex polygon $W_1 W_2 ... W_m$ can be expressed as an analytical inequality via the triangle-area criterion [7]:

$$S_{\Delta QW_m W_1} + \sum_{l=1}^{m-1} S_{\Delta QW_l W_{l+1}} > S_{\square W_1 W_2 ... W_m}, \tag{33}$$

where $S_\Delta$ denotes the triangle area, and $S_\square$ denotes the area of polygon $W_1 W_2 ... W_m$. Suppose that the coordinate of $W_l$ is $(x_{Wl}, y_{Wl})$, each $S_\Delta$ is computed according to

$$S_{\Delta QW_l W_{l+1}} = 0.5 \cdot | x_Q y_{Wl} + x_{Wl} y_{Wl+1} + x_{Wl+1} y_Q - x_Q y_{Wl+1} - x_{Wl} y_Q - x_{Wl+1} y_{Wl} |. \tag{34}$$

$S_{\square W_1 W_2 ... W_m}$ is a constant calculated by summing up multiple triangle areas offline. The aforementioned conditions 1) and 2) are summarized into the following inequalities using (34):

$$S_{\Delta Q V_{jN_j} V_{j1}} + \sum_{l=1}^{N_j - 1} S_{\Delta Q V_{jl} V_{j(l+1)}} > S_{\square V_{j1} V_{j2} ... V_{jN_j}}, \tag{35a}$$

$$\forall Q \in \{A', B', C', D'\}.$$

$$S_{\Delta Q A' B'} + S_{\Delta Q B' C'} + S_{\Delta Q C' D'} + S_{\Delta Q D' A'} > S_{\square A' B' C' D'}, \tag{35b}$$

$$\forall Q \in \{V_{j1}, V_{j2}, ..., V_{jN_j}\}.$$

In (35b), $S_{\square A'B'C'D'}$ denotes the area of the embodied box, thus

$$S_{\square A'B'C'D'} = \left(e_{\text{up}}(t_k) + e_{\text{down}}(t_k) + L_R + L_F\right) \times \left(e_{\text{left}}(t_k) + e_{\text{right}}(t_k) + L_B\right). \tag{35c}$$

Applying (35) to all $k=1,...,N_{fe}-1$ and $j=1,...,N_{obs}$ yields the complete collision-avoidance constraints. By enforcing the ego vehicle's ($N_{fe}-1$) embodied footprints to be collision-free, the ego vehicle's actual footprint will definitely be safe at any a moment throughout $t \in [0,T]$.

For brevity, the collision-avoidance constraints between the embodied footprint $A'B'C'D'$ at $t_k$ and environmental obstacles are abstracted as $\text{EmbodiedFootprints}(t_k) \leq 0$.

## IV. TRAJECTORY PLANNING WITH EMBODIED FOOTPRINTS

This section introduces a trajectory planner based on numerical optimal control, embedded with the embodied footprints proposed in the preceding section. We begin with an overall algorithm architecture in Section IV.A before entering into the detailed functions.

### A. Overall Framework

The steps in the proposed trajectory planner are summarized in the following pseudo-code.

**Alg. 1 Optimization-based Trajectory Planning with Embodied Footprints**

**Function** $traj \leftarrow \text{PlanTrajectory}(task, map)$

1. $path_{\text{coarse}} \leftarrow \text{PlanCoarsePath}(task, map)$;
2. $[traj_{\text{coarse}}, N_{fe}] \leftarrow \text{AttachVelocity}(path_{\text{coarse}})$;
3. $ig \leftarrow \text{ConvertTrajToInitialGuess}(traj_{\text{coarse}}, N_{fe})$;
4. $NLP \leftarrow \text{BuildNLP}(task, map, N_{fe})$;
5. $sol \leftarrow \text{SolveNLP}(NLP, ig)$;
6. $traj \leftarrow \text{ConvertSolutionToTraj}(sol)$;
7. **return**;

In Line 1 of Alg. 1, the function PlanCoarsePath() is used to search a coarse path that connects the initial and goal poses. The inputs of PlanCoarsePath() include *task* and *map*: *task* refers to the initial and goal configurations while *map* presents the environmental layout and obstacle location information. The output of PlanCoarsePath() is a coarse path $path_{\text{coarse}}$ presented by a sequence of waypoints.

In Line 2 of Alg. 1, AttachVelocity() attaches a time course along $path_{\text{coarse}}$ to build a coarse trajectory $traj_{\text{coarse}}$. It is worth emphasizing that AttachVelocity() also decides $N_{fe}$, i.e. the number of collocation points. This is a significant highlight and innovation of this study because our method provides a quantitative estimation of the collocation point scale before solving the NLP problem, which means we resolve the non-collocation error issue without any need to deploy a trial-and-error strategy. Concrete principle of the function AttachVelocity() is introduced in Section IV.B.

In Line 3, the coarse trajectory is converted to an initial guess *ig*, which contains all state and control variables in their discretized forms.

An NLP is built via BuildNLP() in Line 4 of Alg. 1. The detailed principles of this function will be introduced in Section IV.B.

After solving the formulated NLP via a gradient-based optimizer in Line 5, we convert the derived solution *sol* to an optimized trajectory *traj*, which is the final output of Alg. 1.

### B. Velocity Generation in Initial Guess

This subsection presents the principle of AttachVelocity(). In the first step, we assign a kinematically feasible and time-optimal velocity profile to the coarse path. This is achieved by solving a one-dimensional OCP via Pontryagin's Maximum Principle. Using this method, we obtain a coarse trajectory preliminarily.

The second step starts from resampling along the derived coarse trajectory densely. The values of state and control profiles, together with the time stamp, are recorded in each densely resampled waypoint. The first waypoint $wp_0$ refers to the one at $t = 0$, i.e., $wp_0.t = 0$. Thereafter, one checks if the second waypoint $wp_1$ satisfies a relaxed version of the prerequisites (23d), (30b), and (31) with

$$\kappa_k = wp_0.\kappa,$$
$$s_k = wp_0.v \cdot (wp_1.t - wp_0.t). \quad (36)$$

Herein, the prerequisites (23d), (30b), and (31) are relaxed via a user-specified slack variable $0 < \lambda < 1$:

$$|\kappa_k| \cdot s_k \leq \lambda \cdot \frac{\pi}{2}, \quad (37a)$$

$$|\kappa_k| \cdot L_F \cdot \tan(s_k \cdot |\kappa_k|) \leq \lambda \cdot \left(1 + \frac{L_B}{2} \cdot |\kappa_k|\right), \quad (37b)$$

$$\left(1 + \frac{L_B}{2} \cdot |\kappa_k|\right) \cdot \tan(s_k \cdot |\kappa_k|) \leq \lambda \cdot (L_R \cdot |\kappa_k|). \quad (37c)$$

The usage of $\lambda$ serves to relax the right sides of inequalities (23d), (30b), and (31). As a result, (37) becomes a stricter version of these prerequisites (the reason to use $\lambda$ is explained later).

Since the waypoints are densely resampled, $wp_1$ may not violate the prerequisites (37). If so, one continues to check $wp_2$, $wp_3$, etc. until a violating waypoint $wp_m$ is found, which means $wp_{k-1}$ is the last valid one ($m > 1$). Let us discard the intermediate waypoints between $wp_0$ and $wp_{m-1}$. After that, we treat $wp_{m-1}$ like $wp_0$ and repeat the aforementioned operations until the last resampled waypoint is reached. The waypoints that survive from the discard operation are no longer equidistant along the time dimension, but they do form a coarse trajectory $traj_{\text{coarse}}$. The number of waypoints in $traj_{\text{coarse}}$ is regarded as the number of collocation points ($N_{fe}+1$).

Before the end of this subsection, the reason for introducing $\lambda$ in (37) is presented. Recall that the criterion to discard resampled waypoints in AttachVelocity() influences the setting of $N_{fe}$. If one uses the nominal criterion (23d), (30b), and (31), then each $(t_{k+1} - t_k)$ is close to its maximum allowable value. This results in the gross waypoint number of the initial guess being nearly minimized. If one sets $N_{fe}$ to such a low value, then the flexibility in the NLP would be insufficient. More specifically, every change made in $\kappa$ or $v$ during the optimization process may need a smaller bound on $(t_{k+1} - t_k)$, which requires a sufficiently large $N_{fe}$. Therefore, setting $N_{fe}$ too small easily causes an NLP solution failure. The introduction of $\lambda \in (0,1)$ in (37) sets a stricter condition than the nominal one, which makes $N_{fe}$ larger and thus brings more flexibility to the NLP solution process.

### C. NLP Formulation

This subsection presents the NLP formulation, which is comprised of a cost function $J$ and the discretized state and control variables located on $(N_{fe} + 1)$ collocation points. The NLP problem is written in the following form:

$$\begin{aligned} & \underset{z_i, u_i, t_i}{\text{minimize}} \quad J(z_i, u_i, t_i), \\ & \text{s.t., vehicle kinematic constraints;} \\ & \quad \text{two-point boundary value constraints;} \\ & \quad \text{embodied-footprint constraints.} \end{aligned} \quad (38)$$

Herein, the vehicle kinematic constraints refer to the discretized version of the differential equations (2) together with bounding constraints (3). For example, the differential equations in (2) are discretized in the following form:

$$z_i = f(z_{i-1}, u_{i-1}, t_{i-1}), \ i = 1,...,N_{fe}. \quad (39)$$

The two-point boundary-value constraints are imposed on the first and last collocation points for the state/control profiles. The remainder of this subsection elaborates on the embodied-footprint constraints, which are used to replace the nominal collision-avoidance constraints $fp(z_i) \subset \Upsilon_{\text{FREE}}$.

The embodied-footprint constraints are only applied to the collocation points indexed from 1 to ($N_{fe}-1$). This is because the collocation points indexed 0 and $N_{fe}$ correspond to the initial and final positions, which are assumed to be free from collisions. It is worth noting that the collocation points are not forced to be equidistant along $[0,T]$ in our formulated NLP. Instead, one only requires the collocation points to be ordered:

$$t_0 = 0, \ t_{N_{fe}} = T, \quad (40a)$$

$$t_{k-1} \leq t_k, \ k = 1,...,N_{fe}. \quad (40b)$$

Additional decision variables $t_k$, $\kappa_k$, $s_k$, $e_{\text{left}k}$, $e_{\text{right}k}$, $e_{\text{up}k}$, $e_{\text{down}k}$, $\hat{x}_{Ak}$, $\hat{x}_{Bk}$, $\hat{x}_{Ck}$, $\hat{x}_{Dk}$, $\hat{y}_{Ak}$, ..., $\hat{y}_{Dk}$ are introduced in the NLP together with constraints (29), (23d), (30b), (31),

$$\kappa_k = 1/L_W \cdot \tan\phi_k, \quad (40c)$$

$$s_k = v_k \cdot (t_{k+1} - t_k), \quad (40d)$$

and

$$\text{EmbodiedFootprints}(t_k) \leq 0. \quad (40e)$$

In summary, the embodied-footprint constraints consist of (23d), (30b), (29), (31), and (40), which enable our method to plan safe trajectories. It is worth noting that the prerequisites (23d), (30b), and (31) must be satisfied, as our definition of embodied footprints would otherwise be incorrect. However, these prerequisites, which we actively incorporate as constraints in the NLP to ensure they are met, should not be viewed as a deficiency in our planner. In other words, the NLP solver will automatically adjust the decision variables to meet these prerequisites, so users need not verify their fulfillment prior to utilizing our planner. In fact, the efficiency of our proposed trajectory planner only hinges on three assumptions:

1) In discretizing the OCP into an NLP, the state and control variables are piecewise constant.

2) The ego vehicle does not go backward during the driving process.

3) The geometrics and kinematics of the ego vehicle satisfy $2L_W > L_B \cdot \tan\phi_{max}$.

## V. EXPERIMENTAL RESULTS AND DISCUSSIONS

Simulations and field tests were conducted to validate the completeness, solution optimality, solution speed, application scope, and closed-loop tracking performance of our proposed trajectory planner.

### A. Simulation Setup

Simulations were performed in MATLAB (R2023a) and executed on an i9-9900 CPU that runs at 2×3.10GHz. IPOPT [29] is applied in the MATLAB+AMPL environment [30] as the NLP solver. The linear solver embedded in IPOPT is chosen as MA27 from the Harwell Subroutine Library (HSL) [31]. Hybrid A* search algorithm [4] is adopted as the sampling-based path planner in the function PlanCoarsePath() of Alg. 1.

TABLE I. PARAMETRIC SETTINGS FOR SIMULATIONS

| Parameter | Description | Setting |
|---|---|---|
| $L_F$ | Front hang length of the ego vehicle | 0.96 m |
| $L_W$ | Wheelbase of the ego vehicle | 2.80 m |
| $L_R$ | Rear hang length of the ego vehicle | 0.929 m |
| $L_B$ | Width of the ego vehicle | 1.942 m |
| $a_{min}$, $a_{max}$ | Lower and upper bounds of $a(t)$ | -0.75 m/s$^2$, 0.75 m/s$^2$ |
| $v_{max}$ | Upper bound of $v(t)$ | 5.0 m/s |
| $\Phi_{max}$ | Upper bound of $|\phi(t)|$ | 0.7 rad |
| $\Omega_{max}$ | Upper bound of $|\omega(t)|$ | 0.5 rad/s |
| $\lambda$ | Slack parameter in (37) | 0.9 |

The OCP's cost function is defined as the completion time of the driving process. Typically, one might define the cost function as a weighted sum of both time efficiency and trajectory smoothness. However, by focusing purely on time optimality in the OCP, planned trajectories move closer to barriers and/or vertices of obstacles, providing a better demonstration of our proposal's collision-avoidance ability. In seeking time optimality, an intuitive idea is to define the cost function as $J = T$, but such a definition makes the Hessian matrix of the cost function too sparse because $T$ is only related to $t_{N_{fe}}$. As a remedy for this issue, we define $J$ quadratically so that it optimizes the gross time while also minimizing the differences between adjacent collocation points:

$$J = \sum_{k=0}^{N_{fe}-1}(t_{k+1} - t_k)^2. \quad (41)$$

A 30m×30m virtual workspace filled with static polygonal obstacles was constructed. Parametric settings are outlined in Table I. Notably, the basic assumption $2L_W > L_B \cdot \tan\phi_{max}$ holds, thereby justifying the use of our proposed planner in this context.

### B. Simulation Results and Discussions

Three distinct simulation cases were established, each characterized by environmental obstacles with sharp vertices. These cases were designed specifically to probe the ability of our planner to successfully avoid obstacles throughout the driving process. Fig. 8 depicts the optimized trajectories together with the footprints. These footprints, densely resampled between adjacent collocation points, do not overlap with the environmental obstacles, validating the efficacy of our proposed trajectory planner.

Within Fig. 8, the collocation points are marked as red dots along the optimized trajectory. Notably, these points are not equally spaced in the time horizon. Concentrating on Case 3, we provide a further depiction of the optimized $\Delta t_k \equiv t_{k+1} - t_k$ alongside their initially guessed values in Fig. 9. As can be seen from this figure, our trajectory planner generates collocation points that are not uniformly distributed along the time dimension. Additionally, the trends observed for the optimized $\Delta t_k$ and their initially guessed values are similar, suggesting the effectiveness of AttachVelocity(). Additionally, the optimized $\{\Delta t_k\}$ exhibits better uniformity, showing the effectiveness of the employed cost function (41).

Case 2 serves as an example in our examination of how the user-specified parameter $\lambda$ influences trajectory planning performance. Fig. 10 displays the optimized $T$ values under various $\lambda$ ranging from 0.65 to 1.00. As we observe, $T$ generally increases with $\lambda$. The rationale for this trend is as follows. Recall that $\lambda$ is utilized in AttachVelocity() to estimate the distribution of collocation points [refer to (37) in Section IV.B] and $N_{fe}$ is also determined then. Consequently, a larger $\lambda$ results in a smaller $N_{fe}$, and vice versa. When $N_{fe}$ is larger, $\Delta t_k$ is generally smaller, bringing the size of the embodied footprint closer to the actual footprint. This makes the proposed planner less conservative and enhances time optimality, which explains why a descending trend is observed in the optimized $T$ as $\lambda$ decreases.

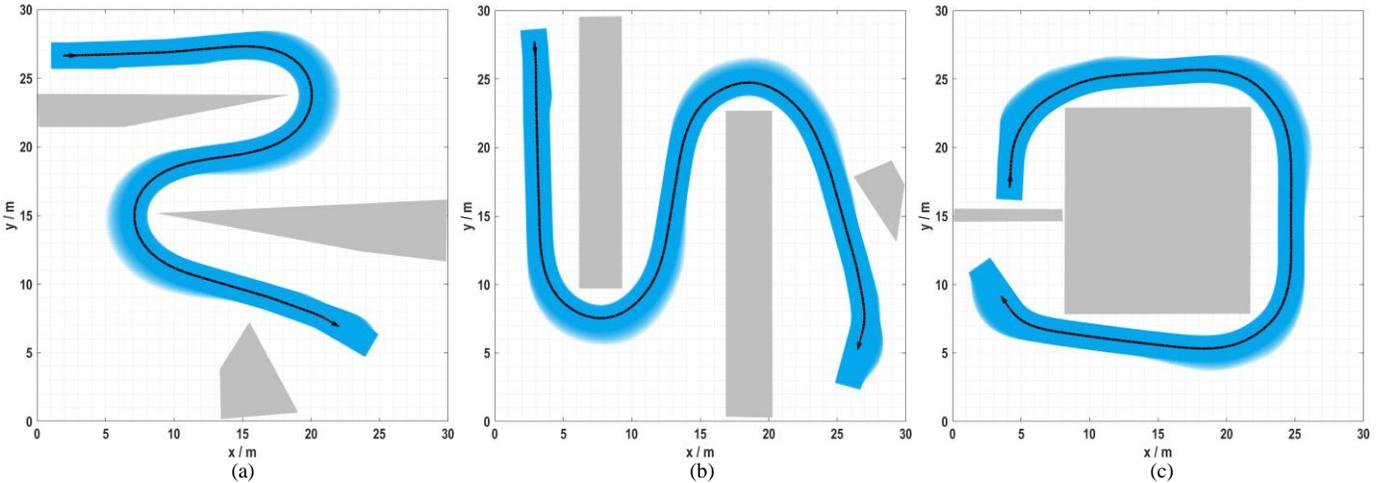

Fig. 8. Footprints associated with optimized trajectories derived by our proposed trajectory planner: (a) Case 1 with $N_{fe} = 117$; (b) Case 2 with $N_{fe} = 124$; (c) Case 3 with $N_{fe} = 105$.

TABLE II. COMPARATIVE SIMULATION RESULTS AMONG FOUR PLANNERS

| Algorithm / Metrics Case ID | Naïve NLP | | | LIOM | | | VPF | | | This work | | |
|---|---|---|---|---|---|---|---|---|---|---|---|---|
| | $T$ (s) | Runtime (s) | Completeness | $T$ (s) | Runtime (s) | Completeness | $T$ (s) | Runtime (s) | Completeness | $T$ (s) | Runtime (s) | Completeness |
| **Case 1** | 17.2297 | 1.531 | √ | 20.0713 | 0.285 | √ | 17.2297 | 1.529 | √ | 17.4909 | 0.757 | √ |
| **Case 2** | 19.5440 | 1.609 | × | 21.7482 | 0.282 | √ | 19.5611 | 4.373 | √ | 19.9812 | 1.749 | √ |
| **Case 3** | 18.7177 | 0.398 | × | 20.0671 | 0.292 | √ | 18.7337 | 1.895 | √ | 19.0464 | 1.005 | √ |

Our proposed trajectory planner is evaluated alongside existing optimization-based trajectory planners, including the naïve NLP method [7], the Lightweight Iterative Optimization Method (LIOM) [1], and VPF [25]. The naïve NLP method is identical to our approach except that the embodied footprint constraints in the NLP (38) are replaced with the nominal collision-avoidance constraints applied to the collocation points. LIOM employs multiple discs to cover the rectangular vehicle body, thus creating buffers to mitigate non-collocation errors. The VPF method was mentioned in Section I.A. When implementing VPF, we strictly follow the pseudo codes and parametric settings in [25]. To ensure fairness in comparison, the number of collocation points in the naïve NLP method, LIOM, or VPF is set to match the value identified by our proposed planner in each case. Thus, the collocation point number remains identical throughout all four planners for each trajectory planning case.

The results of the comparative simulations are summarized in Table II. Here, "completeness" refers to whether a planned trajectory can successfully mitigate non-collocation errors. As shown in the table, the naïve NLP method fails to achieve completeness in two cases. Using Case 2 as an example, Fig. 11 highlights minor collisions between adjacent collocation points, thus showing the limitation of the naïve NLP method and reaffirming our motivation for this research. Although LIOM operates swiftly, it produces overly conservative trajectories, as reflected by the optimized $T$ in each case. VPF is less conservative because its optimized $T$ closely matches that of the naïve NLP method. However, VPF runs slowly as it seeks to determine a valid enlarged box around the ego vehicle's actual footprint in a trial-and-error mode by solving intermediate NLPs iteratively. We showcase the trajectories optimized by the four planners in Fig. 12 using Cases 2 and 3 as examples, which are in line with our aforementioned analysis.

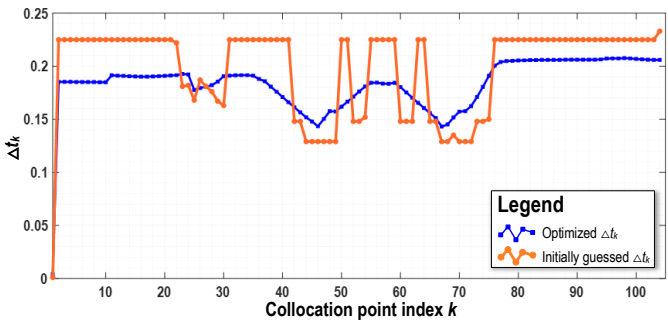

Fig. 9. Comparison between initially guessed and optimized time subinterval durations $\{\Delta t_k | \Delta t_k \equiv t_{k+1} - t_k\}$ in Case 3.

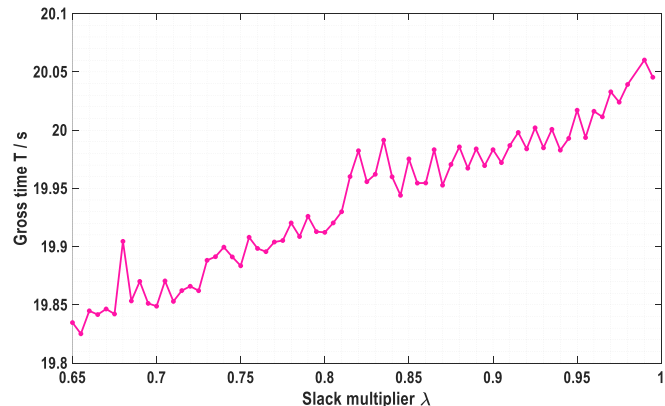

Fig. 10. Migration of optimized $T$ with user-specified slack multiplier $\lambda$ in Case 2.

Upon examining Fig. 12, some readers might argue that VPF offers better solution optimality than our planner because VPF is less conservative. To respond to this concern, we define a new case by adding several static obstacles to Case 3, thus making the environment tighter. Fig. 13a shows the optimized trajectory along with the densely resampled footprints from our proposed planner, whereas VPF, LIOM, and the naïve NLP method fail in that case. Recall that LIOM deploys multiple discs to model the ego vehicle's rectangular footprint. This would render a too large buffer region around the ego vehicle to hinder a narrow passage traverse. That is why LIOM did not work in tiny scenarios. Conversely, the naïve NLP method exits with an optimal trajectory successfully, but the derived trajectory is found to be invalid because it involves non-collocation errors. The failure of VPF is explained as follows. VPF employs a trial-and-error strategy, which starts with solving the naïve NLP problem. If the derived trajectory leads to collisions between adjacent collocation points, then the ego vehicle's footprint is laterally expanded quantitatively in hopes of reducing collisions in the next iteration with this enhanced buffer. In VPF, a variable $\alpha_{gcf} \geq 1.0$ is deployed to describe the quotient of the laterally expanded width divided by the nominal width. According to the monotony policy listed in [25], VPF exits once the calculated $\alpha_{gcf}$ is found to be smaller than that derived in a preceding cycle. In our simulation, $\alpha_{gcf}$ used to be 1.0016 and later got updated to 1.0012, which made VPF exited with a derived optimal solution. However, we find that the derived trajectory involves non-collocation errors, thus the output of VPF is invalid (Fig. 13b). Even without such a monotony policy, VPF is still inefficient to deal with narrow passages when a passage width happens to be smaller than the laterally expanded buffer. At this point, VPF lacks enough flexibility to fight against non-collocation errors. By contrast, our planner does not suffer from such a limitation.

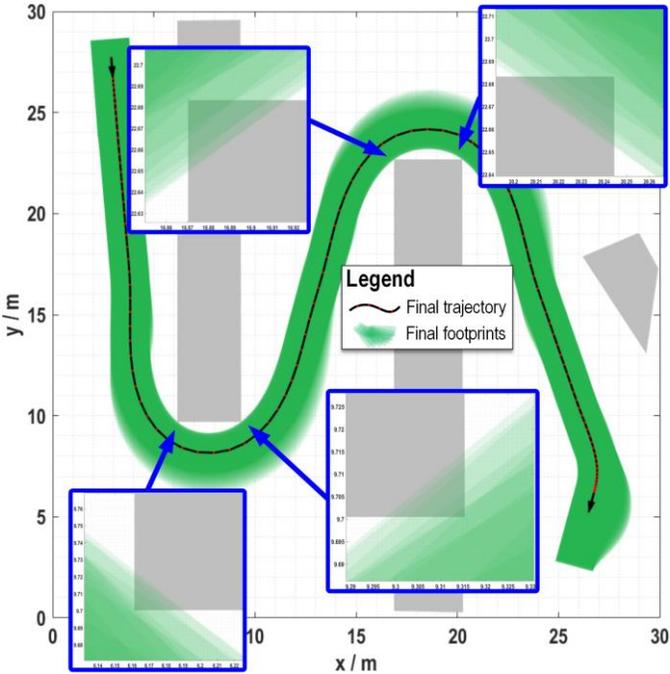

Fig. 11. Footprints associated with an optimized trajectory that is derived by naïve NLP planner in Case 2. Note that collocation points distribute uniformly along time horizon.

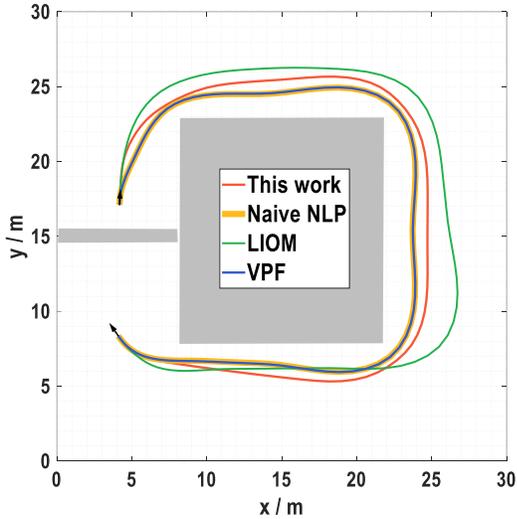

Fig. 12. Comparison among trajectories optimized by four planners.

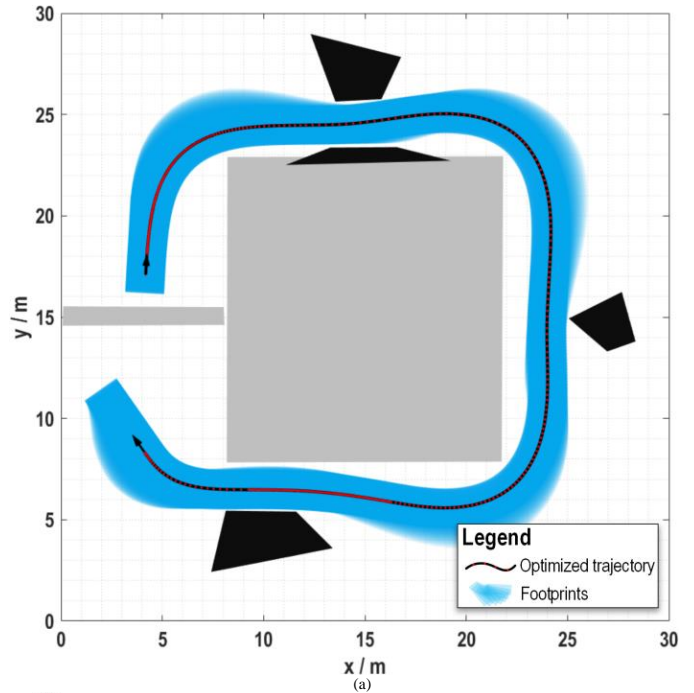

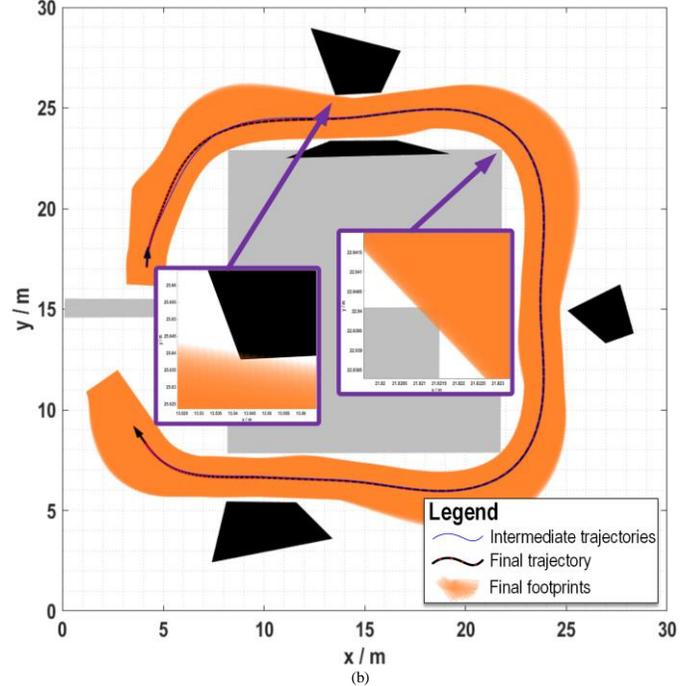

Fig. 13. Comparison between this work and VPF in an enhanced version of Case 3. Note that the black obstacles are newly added to make this new case distinct from Case 3. (a) Footprints associated with an optimized trajectory derived by our proposed planner; (b) Footprints associated with an output trajectory (invalid) derived by VPF.

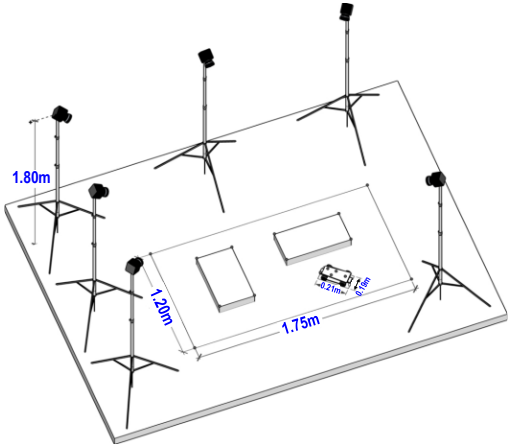

Fig. 14. Workspace layout and indoor localization solution for field tests.

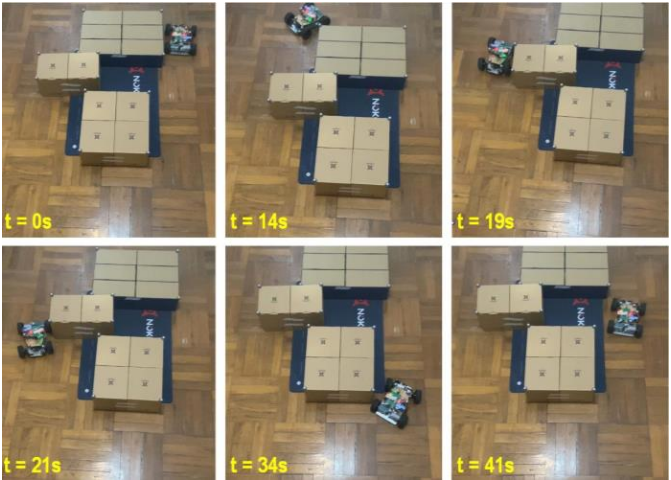

Fig. 15. Snapshots of closed-loop control performance when the ego vehicle is tracking a planned open-loop trajectory, which is derived by a trajectory planner with safe collision-avoidance constraints.

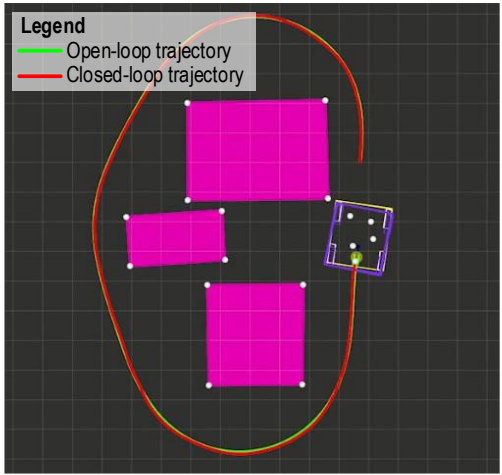

Fig. 16. Comparison between open-loop and closed-loop trajectories in field test.

Fig. 13a plots the distribution of collocation points along the optimized trajectory, indicating that the collocation points are densely packed near the bottleneck regions. This is not determined by users but by the NLP solution process itself. In other words, our planner knows how to distribute collocation points to navigate narrow passages safely. If the passages are really tiny, the NLP solution process knows to slow down with slim steering angles there so as to make the embodied footprints close to the actual footprints there. This unique advantage of our proposed planner clearly demonstrates that tackling narrow passages in a trial-and-error mode is not a complete solution.

The results of the simulation cases discussed in this subsection can be found at bilibili.com/video/BV1J94y1W7JY/. In these simulations, the ego vehicle is bordered in green and the embodied footprint is marked in red.

### C. Field Test Setup, Results, and Discussions

Field tests were conducted on a small-sized autonomous vehicle platform within a 1.75m×1.20m indoor workspace, as shown in Fig. 14. For the purpose of obstacle detection and localization, six infrared sensors were deployed. Specifically, reflective markers were affixed to the vertices of each polygonal obstacle as well as the top of the self-driving vehicle. The procedure involves infrared light sources emitting beams, which are then captured by the infrared sensors after reflection from the markers. This setup allows for the precise localization of each marker [32].

The indoor infrared solution, provided by NOKOV®, was implemented in this study. The infrared sensors from this solution operated at a frequency of 60Hz. The data collected for perception and localization was processed on a desktop computer. This is where the trajectory planning codes, written in C++, were compiled and executed. The trajectory planning module ran once to generate an offline trajectory before the ego vehicle initiated its movement. The precalculated trajectory, along with the localization information, was transmitted from the computer to the autonomous vehicle platform via ZigBee. In regards to onboard tracking control, a Proportional-Integral-Derivative controller was employed for longitudinal tracking, while a Pure Pursuit controller was used for lateral tracking. Each controller was set to a frequency of 10 Hz.

Parameters relevant to the vehicle's kinematics and geometry are detailed in Table III. These parameters satisfy $2L_W > L_B \cdot \tan\phi_{max}$, thus confirming the suitability of the proposed planner for the field tests.

Representative results of the field tests can be accessed via the video link provided in Section V.B. Fig. 15 showcases snapshots of the closed-loop tracking performance, signifying that the planned open-loop trajectory is safe and easily trackable, as further illustrated in Fig. 16.

TABLE III. PARAMETRIC SETTINGS FOR FIELD TESTS

| Parameter | Description | Setting |
|---|---|---|
| $L_F$ | Front hang length of the ego vehicle | 0.036 m |
| $L_W$ | Wheelbase of the ego vehicle | 0.143 m |
| $L_R$ | Rear hang length of the ego vehicle | 0.032 m |
| $L_B$ | Width of the ego vehicle | 0.191 m |
| $a_{min}$, $a_{max}$ | Lower and upper bounds of $a(t)$ | -0.02 m/s², 0.02 m/s² |
| $v_{max}$ | Upper bound of $v(t)$ | 0.25 m/s |
| $\Phi_{max}$ | Upper bound of $|\phi(t)|$ | 0.38 rad |
| $\Omega_{max}$ | Upper bound of $|\omega(t)|$ | 0.10 rad/s |

### VI. CONCLUSIONS

This paper has proposed a theoretical model for safe collision-avoidance constraints for numerical optimal control-based trajectory planners. We have named the proposed model *embodied footprint*, drawing inspiration from the emerging concept *embodied intelligence* from the AI research field. We aim to capture the idea that the footprint model of the ego vehicle proactively knows how to flexibly and autonomously set its buffer scale rather than reactively making adjustments in a feedback or trial-and-error manner.

The proposed method can be further extended to deal with precision-aware motion planning problems in other robotics-related fields other than autonomous driving. The proposed modeling method may also work as a fast collision checker that only needs to examine a small number of waypoints along a to-be-checked path or trajectory.

Our future work is to further extend the current study so that reverse driving conditions are enabled, which are simply symmetric to the analyses in this paper, though.

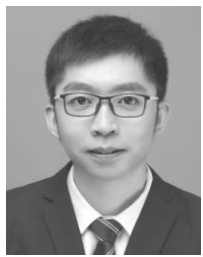

**Bai Li** (SM'13–M'19) received his B.S. degree in 2013 from Beihang University, China, and his Ph.D. degree in 2018 from Zhejiang University, China. From Nov. 2016 to June 2017, he visited the University of Michigan (Ann Arbor), USA as a joint training Ph.D. student. He is currently an associate professor at the College of Mechanical and Vehicle Engineering, Hunan University, China. Before teaching in Hunan University, he worked at JDX R&D Center of Automated Driving, JD Inc., China from 2018 to 2020 as an algorithm engineer. Prof. Li has been the first author of more than 80 journal/conference papers and two books in numerical optimization, motion planning, and robotics. He was a recipient of the International Federation of Automatic Control (IFAC) 2014–2016 Best Journal Paper Prize from Engineering Applications of Artificial Intelligence. He received the 2022 Best Associate Editor Award of IEEE TRANSACTIONS ON INTELLIGENT VEHICLES. He is currently an Associate Editor for IEEE TRANSACTIONS ON INTELLIGENT VEHICLES. His research interest is optimization-based motion planning for autonomous driving.

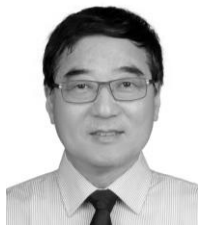

**Youmin Zhang** (M'99–SM'07–F'23) is a Professor at the Department of Mechanical, Industrial, and Aerospace Engineering, Concordia University, Canada. His research interests are in the areas of monitoring, diagnosis and physical fault/cyber-attack tolerant/resilient control, guidance, navigation and control of unmanned systems and smart grids, with applications to forest fires and smart cities in the framework of cyber-physical systems by combining with remote sensing techniques. He has published 8 books, over 600 journal and conference papers. Dr. Zhang is a Fellow of CSME, a Senior Member of AIAA, President of International Society of Intelligent Unmanned Systems (ISIUS) during 2019–2022, and a technical committee member of several scientific societies. He has been an Editor-in-Chief (EIC), Editorial Advisory Board Member of several journals, including as a Member of Board Member of Governors and Representatives for *Journal of Intelligent & Robotic Systems*, Associate Editor for IEEE TRANSACTIONS ON INDUSTRIAL ELECTRONICS, IEEE TRANSACTIONS ON NEURAL NETWORKS & LEARNING SYSTEMS, IEEE TRANSACTIONS ON CIRCUITS AND SYSTEMS - II: EXPRESS BRIEFS, *IET Cyber-systems and Robotics*, *Unmanned Systems*, *Security and Safety*, and Deputy EIC for *Guidance, Navigation, and Control*.

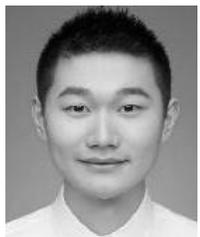

**Tantan Zhang** received the B.S. degree in 2012 from Hunan University, Changsha, China, the double M.S. degrees in 2015 from both Politecnico di Torino, Turin, Italy, and Tongji University, China, and a Ph.D. degree in 2020 from Politecnico di Torino, Italy. He is currently an assistant professor in the College of Mechanical and Vehicle Engineering, Hunan University, China. His research interest is motion planning of automated vehicles.

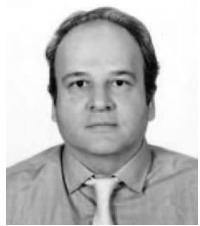

**Tankut Acarman** (M'06) received a Ph.D. degree in electrical and computer engineering from Ohio State University, Columbus, OH, USA, in 2002. He is a Professor and the Head of the Department of Computer Engineering, Galatasaray University, Istanbul, Turkey. He is a co-author of the book entitled *Autonomous Ground Vehicles*. He serves as a Senior Editor for IEEE TRANSACTIONS ON INTELLIGENT VEHICLES. His research interests include aspects of intelligent vehicle technologies, driver assistance systems, and performance evaluation of inter-vehicle communication.

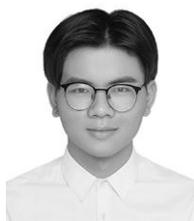

**Yakun Ouyang** (SM'20) received his B.S. degree in 2020 from Nanchang University, China, and his Master's degree in June 2023 from Hunan University, China. He was the first-prize recipient of the 2019 National University Students Intelligent Car Race in China. His research interests include trajectory planning, control, and software engineering of autonomous vehicle systems.

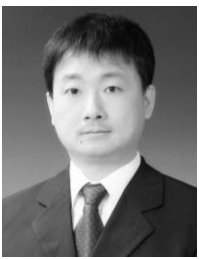

**Li Li** (F'17) is currently a professor with the Department of Automation, Tsinghua University, Beijing, China, working in the fields of artificial intelligence, complex systems, intelligent transportation systems, and intelligent vehicles. He has published over 100 SCI-indexed international journal articles and over 70 international conference papers as a first/corresponding author. He is a member of the Editorial Advisory Board for the Transportation Research Part C: Emerging Technologies, and a member of the Editorial Board for the Transport Reviews and ACTA Automatica. He serves as an Associate Editor of IEEE TRANSACTIONS ON INTELLIGENT TRANSPORTATION SYSTEMS and IEEE TRANSACTIONS ON INTELLIGENT VEHICLES.

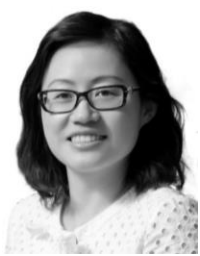

**Hairong Dong** (M'12–SM'12) received the Ph.D. degree from Peking University in 2002. She was a Visiting Scholar with the University of Southampton in 2006 and The University of Hong Kong in 2008. She was also a Visiting Professor with the KTH Royal Institute of Technology in 2011. She is currently a Professor with the State Key Laboratory of Rail Traffic Control and Safety, Beijing Jiaotong University, China. Her research interests include intelligent transportation systems, automatic train operation, intelligent dispatching, and complex network applications. She is a fellow of the Chinese Automation Congress and the co-chair of Technical Committee on Railroad Systems and Applications of the IEEE Intelligent Transportation Systems Society. She serves as an Associate Editor for IEEE TRANSACTIONS ON INTELLIGENT

TRANSPORTATION SYSTEMS, IEEE TRANSACTIONS ON INTELLIGENT VEHICLES, IEEE INTELLIGENT TRANSPORTATION SYSTEMS MAGAZINE, and *Journal of Intelligent and Robotic Systems*.

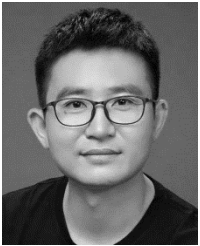

**Dongpu Cao** received the Ph.D. degree from Concordia University, Canada, in 2008. He is a Professor at Tsinghua University, China. His current research interests include driver cognition, automated driving, and cognitive autonomous driving. He has contributed more than 200 papers and 3 books. He received the SAE Arch T. Colwell Merit Award in 2012, IEEE VTS 2020 Best Vehicular Electronics Paper Award, and six Best Paper Awards from international conferences. Prof. Cao has served as Deputy Editor-in-Chief for IET Intelligent Transport Systems Journal, and an Associate Editor for IEEE TRANSACTIONS ON VEHICULAR TECHNOLOGY, IEEE TRANSACTIONS ON INTELLIGENT TRANSPORTATION SYSTEMS, IEEE/ASME TRANSACTIONS ON MECHATRONICS, IEEE TRANSACTIONS ON INDUSTRIAL ELECTRONICS, IEEE/CAA JOURNAL OF AUTOMATICA SINICA, IEEE TRANSACTIONS ON COMPUTATIONAL SOCIAL SYSTEMS, and *ASME Journal of Dynamic Systems, Measurement, and Control*. Prof. Cao is an IEEE VTS Distinguished Lecturer.